\documentclass[letterpaper]{article} 
\usepackage{aaai2026}  
\usepackage{times}  
\usepackage{helvet}  
\usepackage{courier}  
\usepackage[hyphens]{url}  
\usepackage{graphicx} 
\usepackage{natbib}  
\usepackage{caption} 
\usepackage{algorithm}
\usepackage{algorithmic}
\usepackage{amsmath}
\usepackage{booktabs} 
\usepackage{multirow}

\usepackage{newfloat}
\usepackage{listings}
\DeclareCaptionStyle{ruled}{labelfont=normalfont,labelsep=colon,strut=off} 
\floatstyle{ruled}
\newfloat{listing}{tb}{lst}{}
\floatname{listing}{Listing}
\title{RS-OOD: A Vision-Language Augmented Framework for Out-of-Distribution Detection in Remote Sensing}
\author{
    Chenhao Wang\textsuperscript{\rm 1,2}\equalcontrib,
    Yingrui Ji\textsuperscript{\rm 1,2}\equalcontrib,
    Yu Meng\textsuperscript{\rm 1},
    Yunjian Zhang\textsuperscript{\rm 3}, 
    Yao Zhu\textsuperscript{\rm 4},
}
\affiliations{

    \textsuperscript{\rm 1} Aerospace Information Research Institute, Chinese Academy of Sciences \\
    \textsuperscript{\rm 2} School of Electronic, Electrical and Communication Engineering, University of Chinese Academy of Sciences \\
    \textsuperscript{\rm 3} Institution of Information Engineering, Chinese Academic of Sciences
    \textsuperscript{\rm 4} Zhejiang University, Hangzhou

}

\usepackage{bibentry}

\begin{document}

\maketitle

\begin{abstract}
Out-of-distribution (OOD) detection represents a critical challenge in remote sensing applications, where reliable identification of novel or anomalous patterns is essential for autonomous monitoring, disaster response, and environmental assessment. Despite remarkable progress in OOD detection for natural images, existing methods and benchmarks remain poorly suited to remote sensing imagery due to data scarcity, complex multi-scale scene structures, and pronounced distribution shifts.
To this end, we propose RS-OOD, a novel framework that leverages remote sensing-specific vision-language modeling to enable robust few-shot OOD detection. Our approach introduces three key innovations: spatial feature enhancement that improved scene discrimination, a dual-prompt alignment mechanism that cross-verifies scene context against fine-grained semantics for spatial-semantic consistency, and a confidence-guided self-training loop that dynamically mines pseudo-labels to expand training data without manual annotation.
RS-OOD consistently outperforms existing methods across multiple remote sensing benchmarks and enables efficient adaptation with minimal labeled data, demonstrating the critical value of spatial-semantic integration.
\end{abstract}


\section{Introduction}
In any automated or semi-automated building profile extraction process, regardless of whether its source is aerial image, satellite image or lidar point cloud, the initially generated vector data inevitably has "digital original sin". These original outlines are essentially direct translations of pixel grids or discrete point clouds rather than ideal geometry in human cognition. Therefore, they are filled with a large amount of redundant information and high-frequency noise, specifically manifested as: on a macroscopic straight wall, there are jagged edges composed of hundreds of tiny line segments; at the corners or microstructures, they are covered with meaningless tiny bumps or depressions caused by missegment of single or several pixels, i.e. "burrs". \cite{gong2023soda} The core mission of the step of local structure optimization is not to carry out advanced geometric morphology correction, but to assume the basic but crucial responsibilities of "data purification" and "information refining". It uses a series of algorithmic operations to strip away the disturbing information introduced by imaging and processing processes in the original data that is independent of the real building geometry, thereby paving the way for a cleaner, simpler, and more reflective of the macrostructure. This process can be subdivided into two key technical links that are complementary and usually performed in sequence: one is the Douglas-Puk algorithm used to simplify the contours and remove redundant vertices; the other is the glitch removal for smoothing boundaries and eliminating local defects.
Our method achieves markedly higher mask IoU for the smallest object instances, validating the effectiveness of our region magnification and edge refinement strategies.

\begin{figure*}[htbp]
    \centering
    \includegraphics[width=0.8\textwidth]{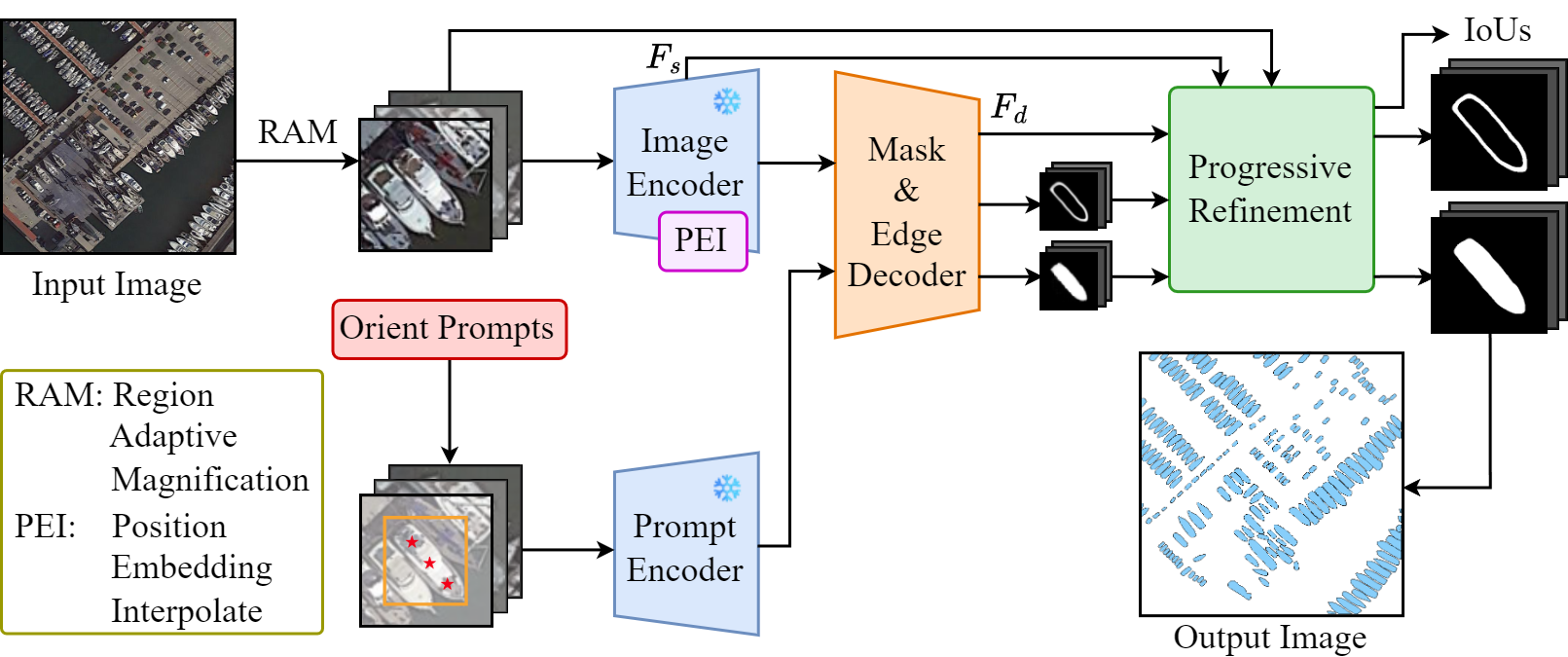}
    \caption{Overview of the  Framework.}
    \label{fig:architecture}
\end{figure*}

\section{Related Work}
Technology deepening: Douglas-Peucker Algorithm (Douglas-Peucker)
Analysis of core issues
What this algorithm needs to solve is a trade-off between "data fidelity" and "data simplicity". A slight jittery wall profile of 500 vertices, although each pixel boundary is "faithfully" recorded in the coordinates, 99\% of the data is redundant to describe the core information of "This is a wall" and greatly increases the burden of subsequent calculations. The essence of the algorithm is that it tries to answer in a non-subjective, quantifiable way: "Of these 500 points, which points are the turning points that define the wall, and which points are mere noise that can be safely removed?"
Selection strategy for parameter 
The soul of the algorithm lies in its only control parameter - the "tolerance" distance. This value is not a random number, it has clear physical and application significance. 
Physical significance: In geographic information or surveying and mapping applications,  represents the "maximum allowable deviation" or "the maximum range of uncertainty for the simplified graph to the original graph". Setting  meters means that you have issued an instruction to the algorithm: "Please simplify this line, but I require that any point on the simplified new line must not exceed half a meter from the original line segment represented by it." This provides a rigid and measurable quality control standard for the simplification process. 
Selection strategy: The value of directly determines the simplified "grain size", and its selection is closely related to the application scenario of the final data.
The challenge is further compounded by the unique characteristics of remote sensing imagery. Small objects often exhibit low contrast against complex backgrounds, suffer from atmospheric effects, and appear at arbitrary orientations. Traditional segmentation networks like Mask R-CNN \cite{he2017mask} show significant performance degradation when directly applied to small remote sensing objects, primarily due to insufficient feature representation at coarse resolutions. This limitation motivates the need for specialized architectures that preserve fine-grained details throughout the processing pipeline.

\noindent
Large scale/high precision applications (such as cadastral surveys, construction details): At this time, every tiny turning point in the building may have legal or engineering significance. So a very small  value is chosen, for example 0.1 meters or less, with the goal of removing only pure pixel-level noise while maximizing all true geometric details. 
Small and medium scale/macro applications (such as urban planning, navigation map): In this scenario, overly detailed outlines are not only unnecessary, but will affect the visual clarity and rendering efficiency of the map. At this time, you can select a relatively large  value, such as between 0.5 meters and 2 meters. This will make the building profile appear in a more general and smoother form, which is in line with macro cognition.
However, several studies have identified limitations when applying SAM to remote sensing imagery, particularly for small objects.  It comprises fine-grained details for segmenting small objects.

\section{Methodology}
To bridge the gap between generic segmentation models and the unique demands of small object segmentation in remote sensing, we propose SOPSeg, a prompt-based framework that introduces three key improvements over SAM: a region-adaptive magnification strategy, an oriented prompt mechanism, and an enhanced decoder with integrated edge prediction. The overall architecture is illustrated in Figure \ref{fig:architecture}.

\subsection{Algorithm execution}
Take a tortuous line composed of 100 vertices $(V_1,\ldots,V_{100})$ extracted from the image, representing the wall on one side of the building, as an example, and set $\varepsilon=0.5$ meters. 
Global Review (First Level Recursion): The algorithm first completely ignores the 98 vertices in the middle and builds a straight reference string between the starting point $V_1 $and the end point $V_{100}. $
Find the point with the largest distance: Then, it will rigorously calculate the vertical distance between each vertex in$ V_2,\ldots,V_{99} and this V_1-V_{100}$ line. Assume that after calculation, it is found that the distance of $V_{47} $is the largest, reaching 1.8 meters. 
Decision: The algorithm compares this maximum distance$ d_{\mathrm{max}}=1.8 $meters with the tolerance we set $\varepsilon=0.5$ meters. Apparently, 1.8>0.5. This result tells the algorithm that $V_{47} $is by no means a slight jitter that can be ignored, it is a structural and critical turning point that must be retained. 
Divide and conquer (to enter the next level of recursion): Since $V_{47} $is "promoted" as the key point, the original problem is divided into two. The algorithm now faces two separate, smaller subproblems: 
Simplify the first curve composed of $V_1,\ldots,V_{47}. $
Simplify the second curve composed of $V_{47},\ldots,V_{100}. $
Recursive termination: This "segment-judgment-resegment" process will continue. Until a subproblem, for example, when simplifying $V_{60},\ldots,V_{75}$, the algorithm found that the distance between all the intermediate points and the reference chords of$ V_{60}-V_{75} $is less than 0.5 meters. At this time, the algorithm will make a final decision: delete all the vertices of $V_{61},\ldots,V_{74},$ and represent them with only a straight line segment from $V_{60} to V_{75}.$
Finally, when all branches have recursion terminated, the set of vertices left behind

\noindent
\textbf{Position Embedding Interpolation.}
Since SAM's positional embedding weights are input-size dependent, the pretrained embeddings trained on $1024 \times 1024$ inputs cannot be directly reused. We address this through bilinear interpolation:

\begin{equation}
    \text{PE}_{\text{target}} = \text{Interpolate}(\text{PE}_{1024}, S_{\text{in}}, S_{\text{in}})
\end{equation}

where $S_{\text{in}} = 256$ for small object processing. This interpolation preserves the relative spatial patterns while adapting to the new resolution.

The combination of region extraction, magnification, and reduced input resolution creates an efficient pipeline: small objects are first magnified to an adequate size, then processed at lower input size without losing critical details.

\subsection{Oriented Prompt Mechanism}

$(V_1,V_{47},V_{60},V_{75},\ldots,V_{100})$ is a high degree of refinement of the original 100 vertices.
1.2. Technology deepening: Burr/Spike Removal
1.2.1. Analysis of core issues
There are many causes of burrs, but they are essentially "misunderstanding" of real geometric forms on the microscopic scale. For example, during the process of raster-to-vector conversion, the discrete step-like boundaries of pixels are mistakenly translated into a series of tiny, alternating internal and external protruding segments. The task of burr elimination is to develop a set of rules with "local geometric insight" so that the algorithm can accurately judge "this is a meaningless noise" rather than "a real architectural detail" like the human eye.
1.2.2. The collaborative logic of criterion combination
A robust burr elimination algorithm must be a set of combined punches, making joint decisions through geometric features of multiple dimensions. 
Sharp angle conditions (morphological diagnosis): A priori knowledge that sharp angles rarely appear in structural angles in artificial buildings. A vertex with an angle less than ${30}^\circ $is highly suspicious in geometry. 
Short-side condition (scaling diagnosis): This condition introduces the concept of scale for the judgment of "sharp angle". If both sides that make up this sharp corner are very short (for example, the length is equivalent to the ground distance represented by a pixel in the source image), then the scale of this feature points to "noise" rather than "design". 

Area Criteria: The area of the triangle composed of three consecutive vertices $P_{i-1}, P_i, P_{i+1} $in the sliding window will approach zero. 
Swivel direction criterion: An outward burr will cause an abnormal reversal of the local vertex direction (clockwise or counterclockwise), destroying topological consistency.
By encoding orientation through point positions rather than explicit rotation parameters, we avoid introducing new learnable components while achieving robust orientation-aware segmentation.

\subsection{Decision making and execution}

The vanilla SAM decoder, while effective for regular objects, often produces blurred edges when dealing with features that have been downsampled.
In actual operation, the system will check this set of combined conditions for each vertex on the outline. Only when a vertex$ P_i $meets multiple high-risk conditions at the same time (for example, its angle is less than$ {30}^\circ,$ and both sides that make it are less than 0.5 meters long), the system will issue a "judgment" on it: remove vertex $P_i $from the vertex list of the outline. After removal, in order to make up for the broken boundary, the system will create a new line segment that directly connects the previous vertex $P_{i-1} $and the next vertex$ P_{i+1} of P_i.$ This "delete and reconnect" operation is like using a scalpel to accurately cut off the tiny flaw and smoothly stitch the wound.
2. Least Squares Fitting
After the local structural optimization of the previous stage, although the building profile has been streamlined in the number of vertices, its essence is still a polyline connected by discrete vertices. Even if a certain section of the wall appears as a straight line in the macroscopic manner, the segments that make it are only the shortest paths connecting the two key vertices that have been retained. Due to the noise of the original data and the nature of the simplified algorithm, these segments are almost impossible to perfectly collinear at the microscope. If these contours are enlarged, you will find that the walls that should be straight show a faint "twist" or "facet" phenomenon, which is unacceptable for applications such as engineering drawings, urban and rural planning, or three-dimensional modeling that require precise geometric definition.
Therefore, the goal of least squares fitting is not just to "straighten the line", but a higher-level "geometric intention inference". It assumes that a series of discrete vertices that form a contour is a noise-free "sampling" of an ideal, perfect geometric primitive (here is a straight line) in the real world. The task of this method is to deduce the unique, mathematically perfect linear equation that is most likely to generate them based on these imperfect sampled data.

\section{Experiments}
\subsection{In-depth analysis of mathematical principles}
\textbf{Datasets.}
Least Squares Method (LSM) has an unshakable position in data science and statistics. Its core idea is to find a model to minimize its "error" or "residue" to observe data. In the two-dimensional straight line fitting scenario, this idea becomes particularly intuitive.
2.1.1. Definition and quantification of errors
For a point set$ {(x_1,y_1),(x_2,y_2),\ldots,(x_n,y_n)} $and a candidate line L, we need a unified criterion to measure how far this set of points "deviates" from this line. This deviation, namely "Error", is the most geometrically intuitive definition of the vertical distance (i.e., orthogonal distance) from each point$ (x_i,y_i)$ to the straight line L, which we denote as $d_i$. The goal of the least squares method is to adjust the position and angle of the straight line L so that the sum of squares of the vertical distances from all points to this line  reaches the minimum value.
2.1.2. The three functions of square 
Uniform error symbol: The distance $d_i $itself has no direction, but the algebraic distance between the point on one side of the line and the point on the other side will be positive and negative. Direct summing will cause positive and negative errors to cancel each other out. The squared operation converts all errors into non-negative values, ensuring that the deviation of each point positively contributes to the total error. 
Robustness: This is the most critical statistical feature of a square term. A point 1 unit away from a straight line contributes to the total error $S 1^2=1$. For a "outlier" with a distance of 3 units, its contribution is $3^2=9$. This disproportionate "punishment" mechanism means that the fitted straight lines will be forced to give priority to those outliers, or in other words, outliers have a greater "voice" to the position and angle of the straight lines. 
Ensure the uniqueness and computability of the solution: Mathematically, the sum of squares of the error, S, is a smooth, continuous convex function about linear parameters such as slope m and intercept c. This means it has only one global minimum. Through calculus, we can find the partial derivative of the linear parameter for S and make it equal to zero, thereby solving an analytical solution (a definite mathematical formula). This ensures that for any set of input points, the least squares method can always give a unique, definite, and optimal linear equation.

\begin{table}[htbp]
\centering
\begin{tabular}{lcc|cc}
\toprule
\multirow{2}{*}{Method} & \multicolumn{2}{c|}{NWPU} & \multicolumn{2}{c}{SAT-MTB} \\
 & IoU & BIoU & IoU & BIoU \\
\midrule
SAM    & 77.80 & 67.71 & 53.68 & 51.82 \\
SAM2   & 77.24 & 65.54 & 52.35 & 48.76 \\
ROS-SAM & 82.84 & 75.09 & 68.43 & 67.26 \\
UGBS    & 86.13 & 79.33 & 70.32 & 69.54 \\
\textbf{SOPSeg}  & \textbf{86.55}	& \textbf{80.49}	& \textbf{73.38}	& \textbf{72.56} \\

\bottomrule
\end{tabular}
\caption{Generalization results on NWPU and SAT-MTB datasets.}
\label{generalize-nwpu-mtb}
\end{table}

\subsection{Ablation Study}
2.1.3. Total Least Squares
In geometric applications, we use strictly speaking the "holistic least squares method" or "orthogonal distance regression". It assumes that both x and y coordinates may have errors, so what is minimized is the vertical (orthogonal) distance from the point to the line. This is different from the more common "Ordinary Least Squares" (OLS). For the fitting of building profiles, using the equation Ax+By+C=0 that expresses straight lines in any direction and minimizes the orthogonal distance is the most rigorous choice in geometry.
2.2. Disassembly of technical implementation process
2.2.1. Step 1: Segmentation
This is the logical starting point of the entire process. The least squares method can only act on the dot sets with "intention expressed as a single segment". 
Data source: The output of Douglas-Pook algorithm naturally completes this segmentation task for us. The simplified vertex sequence output by the algorithm, such as $P_1, P_2, P_3,\ldots, P_k,$ where each pair of adjacent vertices $(P_i, P_{i+1})$ defines an independent fitting unit. 
Key details: What is really used for fitting is not the two points$ P_i $and$ P_{i+1}$, but in the original contour, all the dense original vertices that are simplified into the entire section between $P_i$ and $P_{i+1}$. The system must preserve the topological mapping relationship between the original vertex and the simplified vertex. For example, if the original vertices $V_{58} to V_{83}$ are ultimately simplified to segments $P_4-P_5$, then the data input used for fitting the calculation is the point set$ {V_{58},V_{59},\ldots,V_{83}}.$
2.2.2. Step 2: Fitting Calculation
For each original vertex set determined in the previous step, the system will perform mathematical calculations to determine the parameters A,B,C of the best fit line Ax+By+C=0. 
Computing Core: 

2.2.3. Step 3: Vertex Projection
After calculating the ideal linear equation, we need to use a set of definitions in this line

\section{Conclusion}
Geometric enhancement aims to simulate the morphological changes of buildings at different perspectives, proportions and spatial positions, and improve the model's ability to adapt to geometric deformation.
Technical Strategy:
1. Random rotation and flip: Rotate the input image at random angles (such as 0-360°) and flip horizontally and vertically. The buildings in remote sensing images do not have a fixed orientation. This strategy ensures that the features learned by the model are independent of the direction and avoids identification bias caused by the large number of training samples at a specific angle.
2. Multi-scale cropping and scaling: Generate training samples with different resolutions through random cropping and scaling. The strategy simulates images at different altitudes or sensor resolutions, allowing the model to robustly detect buildings at both large scales (such as large factories, venues) and small scales (such as rural independent houses).
3. Affine transformation and projection perturbation: introduce random affine transformations such as shear, translation, and slight perspective transformation. This can effectively simulate the geometric distortion of buildings caused by sensor tilt shooting or terrain undulations, enhancing the model's contour extraction accuracy in non-orthograms or complex terrain areas.
Compared with conventional data enhancement, the "projection disturbance" introduced by this system can more realistically simulate the perspective distortion introduced by side view imaging, which is crucial for the contour extraction of dense areas of high-rise buildings and can effectively reduce boundary positioning errors caused by interference in the side facade of the building.
(2) Spectral/Radiometric Augmentation (Spectral/Radiometric Augmentation)
Spectral/radiation enhancement is mainly used to deal with the difference in image tone and brightness caused by different sensors, seasons, lighting and atmospheric conditions, and improve the color adaptability of the model.





\bibliography{aaai2026}

\end{document}


\maketitle
\section*{Appendix A: Details of the ReSOS Dataset}
\subsection*{A.1 Dataset Construction Process}
We first cropped the original SODA-A images into 512×512 patches. Oriented bounding boxes from SODA-A were then used as box prompts for our SOPSeg model to generate instance masks. For quality control, we used the model-predicted IoU scores to assess the confidence of each instance prediction. An image patch was discarded if the lowest predicted IoU among all instances in that patch fell below a predefined threshold, set to 0.55 for airplane and helicopter, and 0.65 for all other categories. Additionally, low-quality masks identified through manual inspection were marked as ignore, following the convention in SODA-A.

\subsection*{A.2 Dataset Statistics}

Table~\ref{tab:resos_stats} summarizes the per-category statistics of the ReSOS dataset. The dataset comprises over 709k annotated instances across eight object categories, with a strong emphasis on small objects. We report both the mean instance area (in pixel squared) and the mean absolute size, defined as the larger dimension (height or width) of the instance's bounding box. Across categories, the mean area remains below 800~px$^2$, confirming the small-object nature of the dataset. Categories such as \textit{Swimming Pool}, \textit{Large Vehicle}, and \textit{Container} have relatively larger average sizes, likely due to their physical scale and less crowded spatial distributions. In contrast, \textit{Small Vehicle} instances are both the most abundant and the smallest in absolute scale (mean max dimension of 17.3 pixels), posing significant challenges for instance segmentation models.

\begin{table}[h]
\centering
\caption{Per-category statistics of the ReSOS dataset. The \textit{Mean Absolute Size} is the average of $\max$(height, width) per instance.}
\label{tab:resos_stats}
\begin{tabularx}{\linewidth}{lXXX}
\toprule
Category & Instance Number & Mean Area (px$^2$) & Mean Absolute Size (px) \\
\midrule
Small Vehicle  & 458,105 & 143.8 & 17.3 \\
Container      & 121,895 & 453.9 & 41.9 \\
Ship           & 60,685  & 277.4 & 25.9 \\
Storage Tank   & 35,163  & 294.6 & 18.3 \\
Swimming Pool  & 29,500  & 729.0 & 37.5 \\
Airplane       & 28,311  & 278.5 & 27.7 \\
Large Vehicle  & 15,947  & 560.0 & 46.4 \\
Helicopter     & 1,421   & 409.4 & 33.2 \\
\bottomrule
\end{tabularx}
\end{table}

\subsection*{A.3 Visualization Examples}
\begin{figure*}[htbp]
    \centering
    \includegraphics[width=\textwidth]{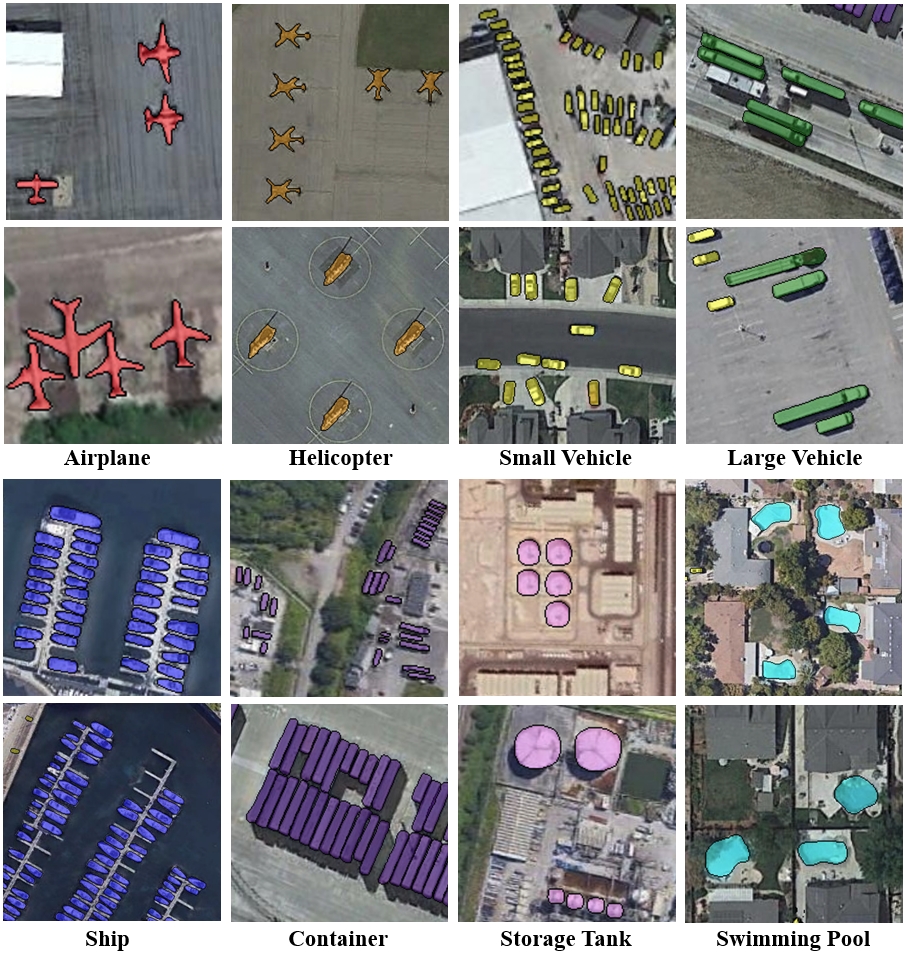}
    \caption{Visualization examples from the ReSOS dataset across eight object categories. Each row shows multiple sample patches with colored instance mask.}
    \label{fig:app1}
\end{figure*}
Fig \ref{fig:app1} show visualization examples from the ReSOS dataset across eight object categories. Each row represents a different object class, with multiple sample patches showing various instances within image crops. The colored masks highlight the precise boundaries of each object instance, demonstrating the dataset's annotation quality and the diversity of object appearances across different scenarios.

\section*{Appendix B: Visualization of Instance Segmentation Results on ReSOS}
\begin{figure*}[htbp]
    \centering
    \includegraphics[width=\textwidth]{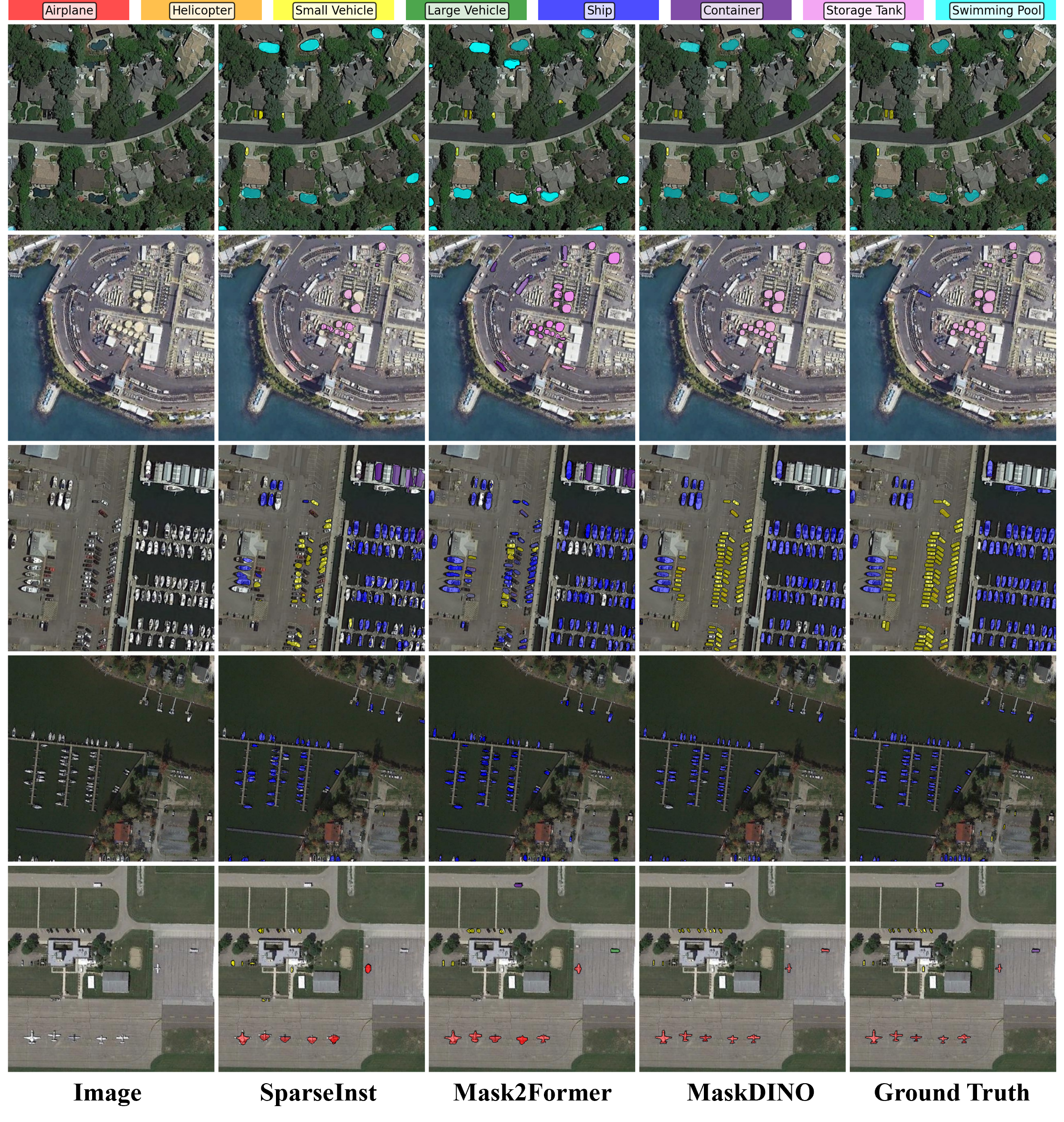}
    \caption{Comparative instance segmentation results on the ReSOS dataset. The ground truth annotations were generated by our SOPSeg.}
    \label{fig:app2}
\end{figure*}
Fig \ref{fig:app2} presents comparative instance segmentation results on the ReSOS dataset. The visualization compares performance across different methods including SparseInst, Mask2Former, MaskDINO, and ground truth annotations, which generated by our SOPSeg. The colored segmentation masks reveal how each method performs on small object instance segmentation. This comparison highlights the varying capabilities of existing methods when handling dense small objects in complex remote sensing environments, particularly for challenging categories like vehicles and aircraft.